# Tracking Multiple Moving Objects Using Unscented Kalman Filtering Techniques

Xi Chen, Xiao Wang and Jianhua Xuan
Bradley Department of Electrical & Computer Engineering
Virginia Polytechnic Institute and State University
900 N. Glebe Road, Arlington, VA 22203, USA
{xichen86, wangxiao, xuan}@vt.edu

The corresponding author: Jianhua Xuan

**Abstract**

It is an important task to reliably detect and track multiple moving objects for video surveillance and monitoring. However, when occlusion occurs in nonlinear motion scenarios, many existing methods often fail to continuously track multiple moving objects of interest. In this paper we propose an effective approach for detection and tracking of multiple moving objects with occlusion. Moving targets are initially detected using a simple yet efficient block matching technique, providing rough location information for multiple object tracking. More accurate location information is then estimated for each moving object by a nonlinear tracking algorithm. Considering the ambiguity caused by the occlusion among multiple moving objects, we apply an unscented Kalman filtering (UKF) technique for reliable object detection and tracking. Different from conventional Kalman filtering (KF), which cannot achieve the optimal estimation in nonlinear tracking scenarios, UKF can be used to track both linear and nonlinear motions due to the unscented transform. Further, it estimates the velocity information for each object to assist to the object detection algorithm, effectively delineating multiple moving objects of occlusion. The experimental results demonstrate that the proposed method can correctly detect and track multiple moving objects with nonlinear motion patterns and occlusions.

## 1. Introduction

It is a challenging task to track multiple moving objects for surveillance and event monitoring, largely due to camera noise, lighting condition, object occlusion, and the varying orientation of different objects. The task of object detection, on the other hand, can be relatively easy to accomplish by many existing techniques such as background subtraction [1] and optical flow techniques [2]. For example, block matching [1, 3, 4] is one of the background subtraction techniques commonly used for moving object detection. The advantage of using block matching in many real applications lies mainly in its fast computational speed as compared to using complicated optical flow techniques. With its low computational complexity, block matching can provide coarse location information for each moving object to be tracked. Tracking of moving objects, based on the results from object detection, is aimed to estimate the optimal trace of the moving objects for further event analysis. If there is only one single moving object, the tracking problem is trivial. However, if there are multiple moving objects sometimes overlapped (or occluded) with each other, the tracking problem become more difficult. When multiple objects depart from each other after occlusion, we particularly need to find and match the correct location for each object, making it possible to detect and track multiple objects all the time.

In this paper, we exploit a simple yet practical idea to help track multiple objects with occlusion. Although the objects overlap with each other, different objects move with different directions and/or speeds, which do not change much during several frames. Considering the continuity of the movement, we can predict the next position of each object based on its estimated velocity when they depart. By comparing to the previously detected objects, we can match the objects after overlapping based on their predicted velocities. Kalman filtering (KF) [5] is widely used to track moving objects, with which we can estimate the velocity and even acceleration of an object with the measurement of its locations. However, the accuracy of KF is dependent on the assumption of linear motion for any object to be tracked. If an object takes some abrupt turns, the nonlinear movement cannot be well handled by the KF framework (due to the linear movement assumption of the design of KF). To solve the nonlinear tracking problem, we propose to develop an unscented Kalman filtering (UKF) technique to track multiple moving objects. UKF [6, 7] has been applied in moving objective tracking [8-11] or dynamic channel tracking [12-14] in wireless signal processing. Therefore, it is a good alternative method for nonlinear motion estimation that is likely required for multiple object tracking in image processing. The UKF method is well designed to utilize unscented transform (UT) [6] to solve nonlinear estimation problems and inhibit the noise impact simultaneously.

The structure of this paper is organized as follows. In Section 2, the block matching algorithm is briefly introduced first for moving object detection. The tracking procedure based on UKF is then detailed with both design and implementation of the tracking algorithm. In Section 3, we report a simulation study for performance evaluation of the approach, and the experimental results on tracking multiple moving objects in real scenarios to demonstrate the effectiveness of the proposed approach. Finally, the conclusion of this paper is summarized in Section 4.

## 2. Method

### 2.1 Moving Object Detection

The goal of block matching [1] is to obtain the motion vector, which is the displacement in block location from the current frame to that in the reference frame. Generally speaking, a block matching method is implemented by the following three components: block determination, searching method, and matching criterion, as shown in Fig. 1. In this paper, we use a three-step search (TSS) method [3, 4] for moving object detection. The TSS method is a fast search technique capable of reducing the number of candidate blocks effectively while at the same time covering a large search area. In a nutshell, the TSS method is purposely designed to direct the search based on the best match from the previous step.

The TSS approach can be briefly summarized as follows:

For each block, performing the following steps iteratively:

**Step 1:** Set the motion vector to zero, and set the best match value to a dissimilarity value of the block in the current frame and the block in the reference frame. If the best match value is less than a predefined threshold, the search stops. If not, go to Step 2.

**Step 2:** The best match is the minimum of the current best match value and the dissimilarity values of the eight neighboring blocks (a step size away centered at current best match location) in the reference frame. A new best match value is determined (or updated) in this step, and the corresponding motion vector is obtained accordingly.

**Step 3:** If the step size is larger than 1, then halve the step size and go to Step 2. If not, return the motion vector.

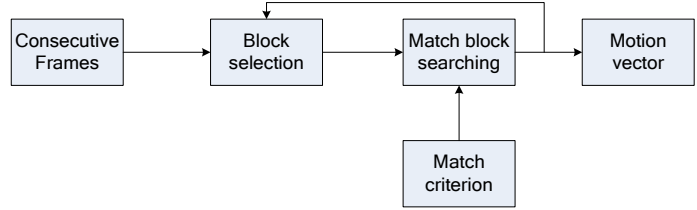

Fig. 1. Flowchart of block matching techniques for moving object detection.

The sum of absolute difference values (SAD) between two blocks is used as the matching criterion in our implementation. Specifically, suppose a block consists of $n \times n$ pixels, for the block $I_c$ in the current frame and the corresponding block $I_r$ in the reference frame (which is shift by a displacement of $(p,q)$), the matching criterion $M(p,q)$ is defined by the following equation:

$$M(p,q) = \sum_{i=1}^{n}\sum_{j=1}^{n}|I_c(i,j) - I_r(i+p, j+q)| \qquad (1)$$

The block with the smallest SAD value is selected as the best match block, and the motion vector can be calculated accordingly.

**2.2 Moving Object Tracking with Kalman Filtering**

Using the fast detection algorithm described above, we can obtain the central positions of moving objects as detected. Due to the noise and limitation of the detection method, such a central position representation sometimes does not reflect the accurate location of a moving object. When multiple objects occlude, the positions need to be further delineated in order to track each object. Through the tracking approach described in this section, we hope to obtain an optimal estimation of the path of each object, hence, to achieve robust tracking of multiple moving objects with occlusion.

**2.2.1 Kalman filtering**

Kalman Filter (KF) was proposed by R. E. Kalman in 1960 [5], which addressed the problem of extracting the useful signal from noisy measurement variables. R. E. Kalman adopted the idea of state space representation and incorporated it into statistical estimation theory for the development of this filtering technique. Based on the statistical characteristics of the system noise and measurement noise, the measurement variables are used as input signal, and the estimation variables that we need to know are the output of the filter. The whole filtering process is composed of a prediction equation and an update equation, which also serve to portray the entire system, as defined by Equation (2):

$$\begin{aligned} X(n) &= F \cdot X(n-1) + V_q(n-1) \quad \text{(prediction equation)} \\ Y(n) &= H \cdot X(n) + V_p(n) \quad \text{(update equation)} \end{aligned} \qquad (2)$$

where $X(n)$ and $Y(n)$ are the estimated state variable and measurement variable, respectively. $F$ is the state transition matrix and $H$ is the measurement matrix. $V_q(n)$ and $V_p(n)$ represent the system noise and measurement noise, respectively.

Although KF is designed based on a criterion of linear minimum square errors, the optimal estimation can be achieved by Kalman filtering only if the following three assumptions are met simultaneously: (1) the state model and measurement model are both linear; (2) the model is designed well to fit the actual system; (3) the noise is additive Gaussian noise. In actual or real systems, the second and third assumptions are quite difficult to be met. The designed model will deviate, more or less, from the actual system. Additionally the noise environment is more complicated in real applications. To overcome these limitations, many improved filtering techniques have been proposed to extend the implementation of KF. The most successful filter, at least until now, is the unscented Kalman filter (UKF), which treats the system from quite a different perspective, however, keeping the basic scheme of KF intact.

**2.2.2 Unscented Kalman filtering**

Unscented Kalman filter (UKF) was proposed by Julier and Simon in 1995 [6]. Since then, the Kalman filter theory has been widely used for signal estimation and trace tracking in many real applications. The core idea of UKF is to address the problem of nonlinear systems and to certain extent, the non-Gaussian noise. Unscented transform (UT) [7] is adopted in UKF to replace the linear transform, thus enabling Kalman filtering to deal with nonlinear systems directly.

For a nonlinear system, assuming that the state variable $X$ is an $N$-dimensional vector, we can obtain $2N+1$ sigma samples $X_i$ and their related weight functions $W_i$ based on the state variables $X$ and its covariance $P$. The criterion for sigma sample generation is based on the unscented

transform as defined by Equations (3)-(10) [7]:

$$X_0 = X, i = 0 \tag{3}$$

$$X_i = X + \sqrt{(N+\lambda)P_i}, i = 1, 2, ..., N \tag{4}$$

$$X_i = X - \sqrt{(N+\lambda)P_{(i-N)}}, i = N+1, N+2, ..., 2N \tag{5}$$

$$W_0^m = \lambda/(N+\lambda) \tag{6}$$

$$W_0^c = \lambda/(N+\lambda) + (3-\alpha^2) \tag{7}$$

$$W_i^m = W_0^c = 1/2(N+\lambda), i = 1, 2, ..., 2N \tag{8}$$

$$w^m = [W_0^m \cdots W_{2N}^m]^T \tag{9}$$

$$W = (I - [w^m \cdots w^m]) \times diag(W_0^c \cdots W_{2N}^c) \times (I - [w^m \cdots w^m])^T \tag{10}$$

where $\lambda = 3\alpha^2 - N$ and parameter $\alpha$ controls the range of the sigma samples around the mean value, usually, $1e-4 \leq \alpha \leq 1$.

The key concept of UKF can be summarized as follows: assuming that the prior probability distribution of the state variable $X$ is subject to Gaussian, with its mean value and its covariance matrix, we can calculate the predicted mean value and covariance matrix using the posterior probability distribution after a nonlinear transform of the system. Assuming that the system dynamic change contains nonlinear variations, prediction function can be represented as $X(n+1) = f(X(n))$ and the observation $Y$ is a function of current state $X$, which is represented as $Y(n) = h(X(n))$. The filtering scheme of UKF is similar to that of KF, as described by a set of equations below:

$$X(n-1) = [x(n-1) \cdots x(n-1)] + \sqrt{3}\alpha \begin{bmatrix} 0 & \sqrt{P(n-1)} & -\sqrt{P(n-1)} \end{bmatrix} \tag{11}$$

$$\hat{X}(n) = f(n-1, X(n-1)) \tag{12}$$

$$\hat{X}(n \mid y_{n-1}) = \hat{X}(n)w^m \tag{13}$$

$$P(n, n-1) = \hat{X}_n W(\hat{X}_n)^T + V_q(n-1) \tag{14}$$

$$X(n) = [\hat{X}(n \mid Y(n-1)) \cdots \hat{X}(n \mid Y(n-1))] + \sqrt{3}\alpha \begin{bmatrix} 0 & \sqrt{P(n,n-1)} & -\sqrt{P(n,n-1)} \end{bmatrix} \tag{15}$$

$$Y(n) = h(n, X(n)) \tag{16}$$

$$K(n) = X(n)WY^T(n)[Y(n)WY^T(n) + V_p(n)]^{-1} \tag{17}$$

$$\hat{X}(n \mid Y(n)) = \hat{X}(n \mid Y(n-1)) + K(n)[Y(n) - Y(n)w^m] \tag{18}$$

$$P(n) = P(n, n-1) - K(n)[Y(n)WY^T(n) + V_p(n)]K^T(n) \tag{19}$$

where $\hat{X}(n \mid Y(n))$ is the current estimation of state variable $X(n)$ under the observation of $Y(n)$, which is the output of UKF as the algorithm converges.

Although both KF and UKF are based on the same scheme, the transform, UT, makes the latter applicable to nonlinear systems. It is worth noting that UKF can deal with all the scenarios that KF is able to, nonetheless, applicable to a much larger field than KF for many nonlinear applications. In addition, UKF can often provide better estimation accuracy than KF can; sometimes, even if the noise is not Gaussian distributed, UKF can still work, while KF may fail, to provide sub-optimal estimation results for state variables.

**2.2.3 Design of system equations for UKF**

The design of prediction and update models is the key aspect of the Kalman filtering theory. In our experiments, the movement of one object in two-dimensional (2-D) image can be seen as a combination of movements in x-axis and y-axis, and these two motion components can be processed independently. Assuming that there are $M$ moving objects detected, $2M$ motion components (i.e., $M$ components in x-axis and $M$ components in y-axis) are needed to be estimated for tracking. Here the measurement vector $Y$ can be expressed as

$$Y = [x_1, y_1, x_2, y_2, ..., x_M, y_M]^T \tag{20}$$

where $\{(x_i, y_i) | i = 1, ..., M\}$ represent the location information for the $i$-th detected object in x-axis and y-axis directions.

For each movement, we want to use a proper equation to reflect the dynamics, so the velocity and acceleration of the object are both considered. The state variable $X$ is thus defined as

$$X = [x_1, v_{x1}, a_{x1}, y_1, v_{y1}, a_{y1}, ..., x_M, v_{xM}, a_{xM}, y_M, v_{yM}, a_{yM}]^T \tag{21}$$

where $\{(v_{xi}, v_{yi}) | i = 1, ..., M\}$ and $\{(a_{xi}, a_{yi}) | i = 1, ..., M\}$ represent the velocity and the acceleration of the $i$-th object, respectively. The moving model for the $i$-th object is therefore defined by a nonlinear system as follows:

$$x_i(n) = x_i(n-1) + v_{xi}(n-1)dt + \frac{1}{2}a_{xi}(n-1)dt^2$$
$$y_i(n) = y_i(n-1) + v_{yi}(n-1)dt + \frac{1}{2}a_{yi}(n-1)dt^2 \tag{22}$$

Accordingly, the transition matrix can be represented as

$$F = \begin{bmatrix} F_1 & 0 & \cdots & & 0 \\ 0 & \ddots & & & \\ \vdots & & F_i & & \vdots \\ & & & \ddots & 0 \\ 0 & & \cdots & 0 & F_M \end{bmatrix} \tag{23}$$

where

$$F_i = \begin{bmatrix} 1 & dt & dt^2/2 & 0 & 0 & 0 \\ 0 & 1 & dt & 0 & 0 & 0 \\ 0 & 0 & 1 & 0 & 0 & 0 \\ 0 & 0 & 0 & 1 & dt & dt^2/2 \\ 0 & 0 & 0 & 0 & 1 & dt \\ 0 & 0 & 0 & 0 & 0 & 1 \end{bmatrix} \quad (i=1, ..., M) \tag{24}$$

Considering the form of measurement variable $Y$, the measurement matrix $H$ is defined as

$$H = \begin{bmatrix} H_1 & 0 & \cdots & & 0 \\ 0 & \ddots & & & \\ \vdots & & H_i & & \vdots \\ & & & \ddots & 0 \\ 0 & & \cdots & 0 & H_M \end{bmatrix} \tag{25}$$

where
$$H_i = \begin{bmatrix} 1 & 0 & 0 & 0 & 0 & 0 \\ 0 & 0 & 0 & 0 & 0 & 0 \\ 0 & 0 & 0 & 0 & 0 & 0 \\ 0 & 0 & 0 & 1 & 0 & 0 \\ 0 & 0 & 0 & 0 & 0 & 0 \\ 0 & 0 & 0 & 0 & 0 & 0 \end{bmatrix} \quad (i=1, \ldots, M) \quad (26)$$

As a result, the UKF scheme, as described in Equations (11)-(19), can be used to estimate the state variable $X$ from the nonlinear system (Equation (22)) for multiple object tracking.

Based on the above description, the tracking algorithm using UKF can be summarized as follows (see Fig. 2 for a flowchart of the algorithm):

**Step1:** According to the results from object detection, we can know how many objects are detected and subsequently need to be tracked. The transition matrix $F$ and measurement matrix $H$ can be initially defined (see Equations (23) and (25)). The initial value of state vector $X$ is obtained accordingly.

**Step2:** Apply the prediction model of UKF to find the next positions and velocities of objects (Equations (13)-(19)).

**Step 3:** Feed the velocity of each object back to the detection algorithm to resolve the ambiguity in object detection when occlusion occurs.

**Step 4:** Update the state vector with the measurement value of each object's location (Equation (18)). Then, go back to Step 2 to begin next iteration.

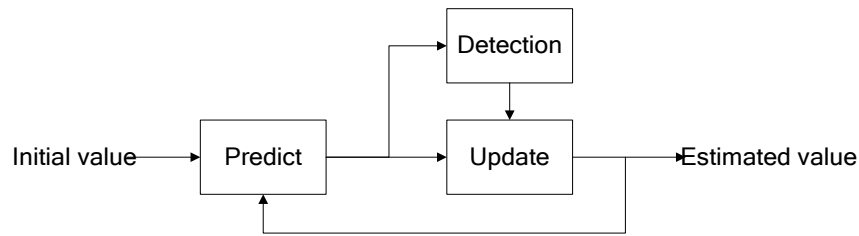

Fig. 2. Flowchart of the proposed motion tracking approach using UKF.

## 3. Experimental Results

**3.1 Simulation Study**

In real applications, since we do not know the ground truth about the movement of moving objects, it is quite hard to compare the performance of UKF and KF directly based on real images. Considering that the main difference between two filters usually occurs when moving objects take turns, we design a turning motion by simulation and compare the tracking performances of both filters (i.e., UKF and KF). Assuming that the errors in the detected results are subject to Gaussian distribution, we add different levels of noise to the original path and then examine the tracking performances of UKF and KF, respectively.

The experimental results are summarized in Table 1 in terms of tracking errors (mean squared errors). From the table, we can see that when the noise level is quite low (e.g., $\sigma=1$), the tracking error of UKF between the estimated and original paths is quite low. As a matter of fact, in this case, it is much lower than the error of KF, which indicates that the nonlinear motion affects the performance of KF more significantly than the noise does. UKF works much better than KF in dealing with the nonlinear motion. As the level of noise increases, the nonlinear motion affects the estimation accuracy of both filters more drastically. Therefore the improvement of UKF over KF degrades as shown in Table 1. Nevertheless, UKF still provides better estimation accuracy than KF does. When the noise level comes to $\sigma=10$, the performances of UKF and KF are relatively close. In Fig. 3, we also show the UKF estimated path compared to the original true path, where the noise level is $\sigma=3$. From the figure, we can see that the tracking result is quite good with the estimated result from UKF.

Table 1. Comparison of estimation errors under different noise conditions

| Noise level ($\sigma$) | 1 | 3 | 5 | 10 |
|---|---|---|---|---|
| Tracking error (UKF) | 0.79 | 1.57 | 2.33 | 4.88 |
| Tracking error (KF) | 1.51 | 2.07 | 2.95 | 5.53 |

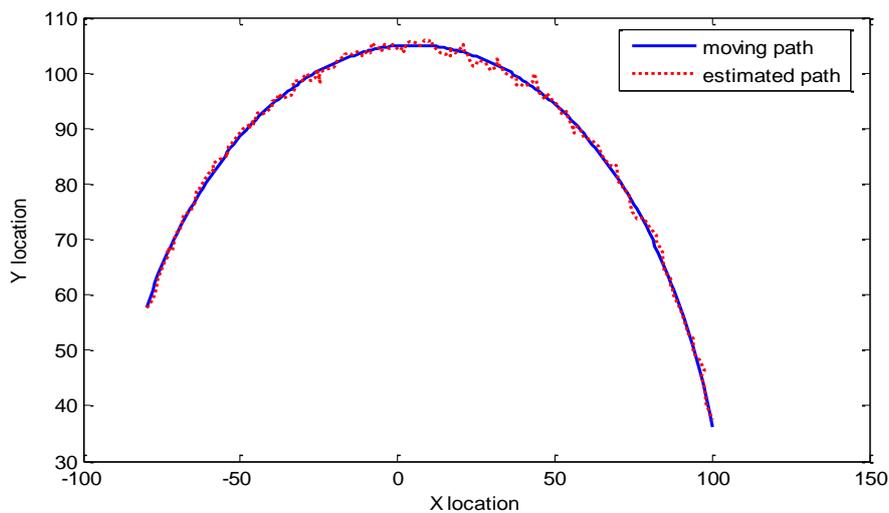

Fig. 3. Tracking performance of UKF in a simulated study with nonlinear motion of turning.

### 3.2 Detection of Moving Objects

We have used the block matching technique described in Section 2 to obtain initial motion vectors from two real image sequences. The procedure of our proposed detection algorithm can be described as follows. The input images are first divided into disjointed blocks with the same fixed size, and the three-step search (TSS) algorithm is then used to search for the best match block with the smallest sum of absolute difference values (SAD). Only the regions containing continuous blocks with non-zero motion vectors are designated as moving objects [15, 16]. This is based on the assumption that the moving objects usually cover multiple blocks in an image sequence, As a result of this assumption, the influence of noise can be alleviated to a great extend.

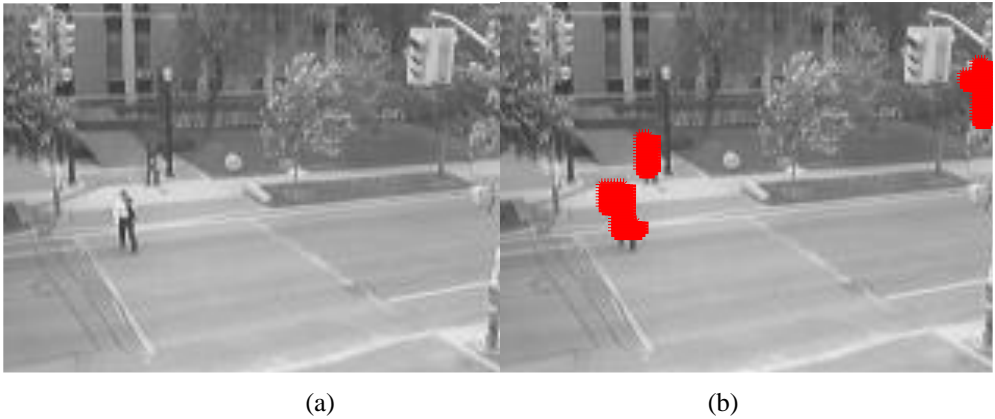

(a)  (b)

Fig. 4. Moving object detection: (a) an exemplary image of original input image sequence; (b) detected moving objects.

Fig. 4 (a) shows one of the two input images, in which two people are walking across the street while the leaves at the top-right corner are also wavering. Fig. 4 (b) indicates the places where the values of motion vectors are non-zero as obtained from the first step of our algorithm. The isolated non-zero motion vectors are mainly caused by noise and slightly moving in the background, such as slightly swaying of some of the leaves. Comparing Fig. 4 (b) with Fig. 4 (a), we can see that our proposed algorithm is quite effective in detecting moving objects. However, the swaying leaves at the top-right corner are erroneously introduced by our detection algorithm. Although there is some error in moving object detection, based on the detection results after multiple frames, we know that these regions do not have any consistent movement and the detection results often fall in a small group of pixels. Such an object is directly deleted from our detection results for further analysis (categorized as 'irrelevant' movement, usually caused by background noise).

### 3.3 Tracking of Multiple Moving Objects

We have implemented the proposed UKF approach and tested the algorithm on several moving objects with complex motion patterns [17, 18]. Due to the space limitation, we only report some exemplary results on multiple object tracking in this section. The complete tracking algorithm can be summarized

as follows:

**Initialization:** Select the first image as the reference image and the second image as the current image. Assuming the initial location of the M moving object are $L_1$, $L_2$, …, $L_M$, respectively. The detection and tracking parts are both initialized.

**Step 1 (Detection):** Calculate motion vector of the current image using a block matching algorithm [5]. For the $i$-th moving object, we search around $L_i$ in the current image for the region with non-zero motion vector values, and then calculate the center of the region as the detected location of the moving object.

**Step 2 (Tracking):** Pass the location point as the measurement value to the UKF update process to obtain the estimated value of current each moving object's location. The UKF estimates or predicts all the objects simultaneously and independently. The prediction process of UKF is then invoked and the predicted velocity is fed back to the detection step (Step 1).

**Step 3:** Assign the current image as the reference image and the next image in the input image sequence as the current image; go to Step 1.

For the tracking of multiple moving objects, the occlusion among objects is our main concern. If the moving objects are just located at different places and move with no overlap, the case is nearly the same as we detect and track multiple independent objects simultaneously. What we want to tackle in this study is the scenarios where moving objects are quite close sometimes, where occlusion exits. As such, we apply our detection and tracking scheme to complicated and challenging cases when there are multiple moving objects with occlusion.

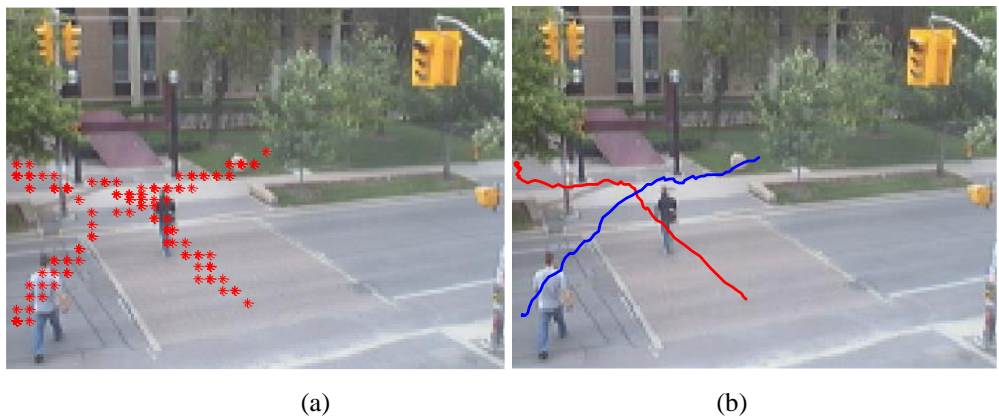

(a)  (b)

Fig. 5. Detection and tracking of multiple moving objects: (a) detected objects (represented in red dots); (b) tracking of two detected objects with occlusion.

Figure 5 shows the detection and tracking results when two moving objects get close and occlusion occurs [10]. We can see that when the two moving object are quite close, the performance of detection degrades. Further, if only based on the detection results (location information), we cannot separate the correct objects after occlusion. Based on the tracking approach, i.e., the UKF approach in our case, we can obtain more accurate location information for each person in motion. By considering the continuity of velocity of a moving object, we can match the detected result after occlusion to that before occlusion. This is quite important to resolve the ambiguity existing in most object detection methods. To provide a more clear idea about this, we also show the velocity information for each person in Fig. 6, the velocity components (in x-axis and y-axis) of each moving object (person). From the figure, we can see that each person's velocity pattern is quite different from the other. Thus, based on such information, we can clearly predict the next position after occlusion and find the correct one around our predicted position from tracking each object continuously.

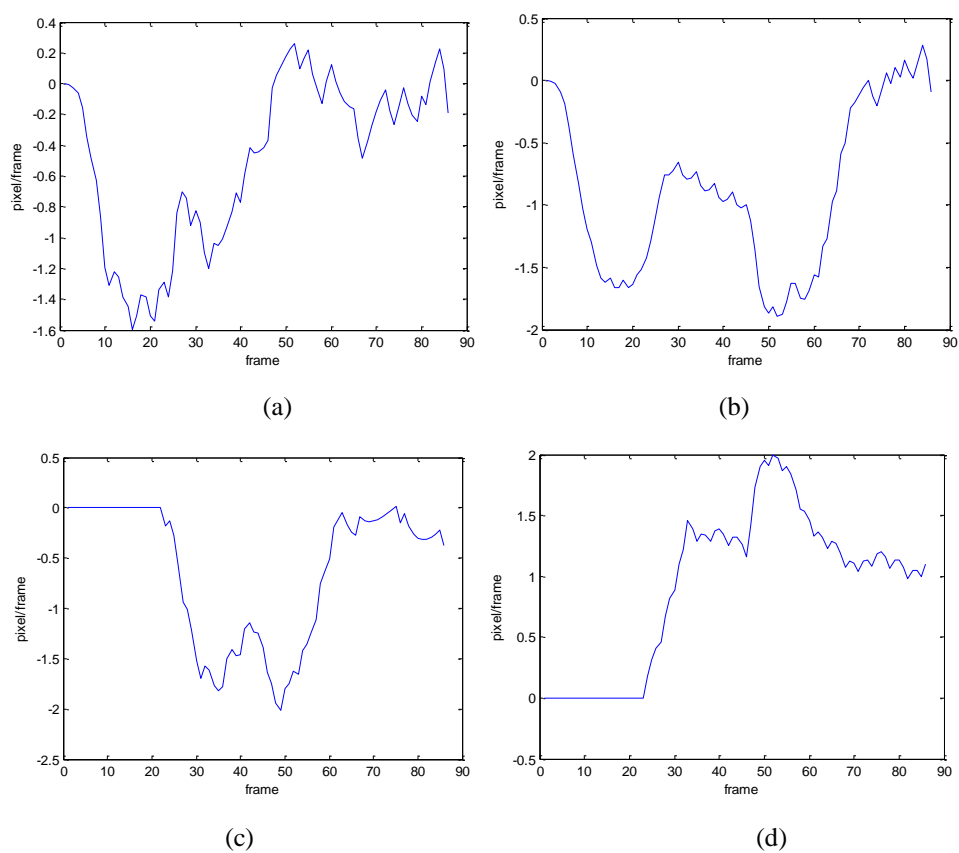

Fig. 6. Estimated velocity information of tracking part: (a) vertical velocity of Object #1; (b) horizontal velocity of Object 31; (c) vertical velocity of Object #2; (d) horizontal velocity of Object #2.

Further, we apply our method to a more challenging case, as shown in Fig. 7 [11]. In this scenario, we have three persons, and each of them would have overlapped with the others during the moving process. The background of the image is quite complicated, which further affects the accuracy of detection. Even if our detection method can identify the moving objects in each image, the ambiguity caused by the occlusion is more challenging. We apply our detection and tracking scheme to this case, and the results are shown in Fig. 8. Assisted by the velocity information estimated by UKF for each person in previous image, the detection algorithm can identify the correct person even with occlusion in current image. Consequently, the detection algorithm provides the location information to the tracking algorithm (i.e., UKF); UKF further improves the estimation accuracy and obtains the final location of each person in the current image. From Fig. 8, we can clearly see that our tracking scheme (as implemented by UKF) can work well in such a complicated case, showing the detailed movement for each person before and after occlusion.

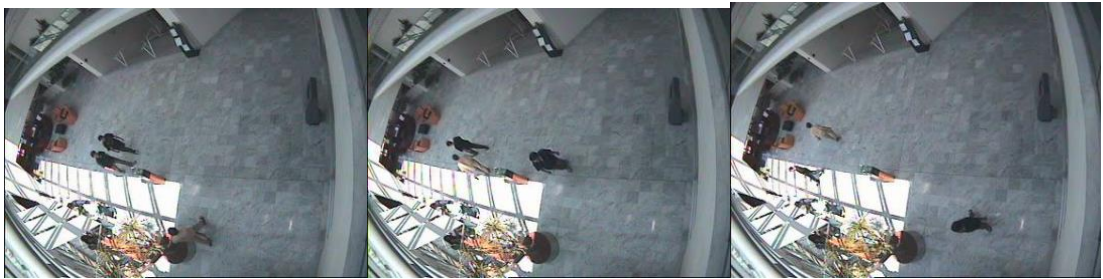

Fig. 7. An image sequence for video surveillance and monitoring; multiple moving objects with occlusion are presented in the sequence.

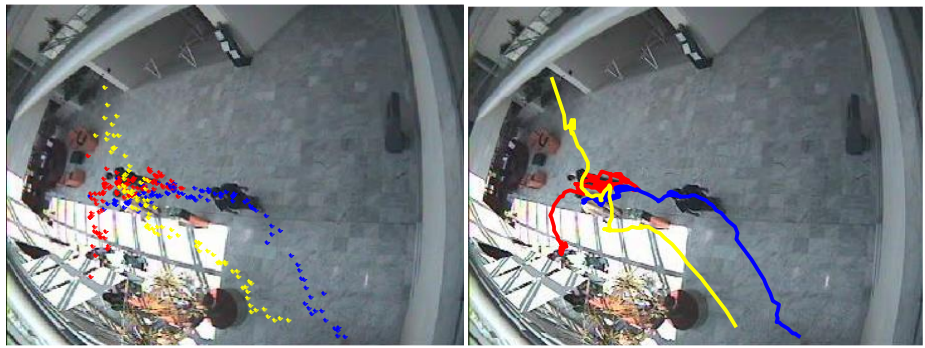

Fig. 8. Final detection and tracking results: (a) detected objects (colored in red, yellow and blue); (b) tracking moving objects with occlusion.

## 4. Conclusion

In this paper, we have proposed and implemented an effective approach for multiple moving object tracking. Specifically, a UKF-based tracking algorithm is developed to optimize the nonlinear moving paths of the objects with occlusion. UKF estimates both the location and velocity information of moving objects, and finally provides more accurate location information. With the feedback of tracking algorithm, especially the velocity information, the continuity of object movement can help the detection algorithm identify the correct objects when occlusion occurs, which to a great extent resolves the ambiguity in detection and tracking of multiple moving objects. The experimental results demonstrate that the proposed method can correctly detect and track multiple moving objects and work well in occlusion scenarios.